\documentclass[10pt,twocolumn,letterpaper]{article}

\usepackage{iccv}              

\usepackage{diagbox}
\usepackage{cases}
\usepackage{algorithmic}
\usepackage{algorithm}
\usepackage{array}
\usepackage{xcolor} 
 
\usepackage{multirow}

\definecolor{iccvblue}{rgb}{0.21,0.49,0.74}
\usepackage[pagebackref,breaklinks,colorlinks,allcolors=iccvblue]{hyperref}

\def\paperID{11903} 
\def\confName{ICCV}
\def\confYear{2025}

\title{Boosting Multimodal Learning via Disentangled Gradient Learning}

\author{Shicai Wei\textsuperscript{1} \quad\quad\quad Chunbo Luo\textsuperscript{2}* \quad\quad\quad  Yang Luo\textsuperscript{2}  \\
\textsuperscript{1}The Laboratory of Intelligent Collaborative Computing of UESTC \\
\textsuperscript{2}The  School of Information and Communication Engineering of UESTC \\
{\tt\small shicaiwei@std.uestc.edu.cn,  \{c.luo, luoyang\}@uestc.edu.cn }
}

\begin{document}
\maketitle
\begin{abstract}
Multimodal learning often encounters the under-optimized problem and may have worse performance than unimodal learning. Existing methods attribute this problem to the imbalanced learning between modalities and rebalance them through gradient modulation. However, they fail to explain why the dominant modality in multimodal models also underperforms that in unimodal learning. In this work, we reveal the optimization conflict between the modality encoder and modality fusion module in multimodal models. Specifically, we prove that the cross-modal fusion in multimodal models decreases the gradient passed back to each modality encoder compared with unimodal models. Consequently, the performance of each modality in the multimodal model is inferior to that in the unimodal model. To this end, we propose a disentangled gradient learning (DGL) framework to decouple the optimization of the modality encoder and modality fusion module in the multimodal model. DGL truncates the gradient back-propagated from the multimodal loss to the modality encoder and replaces it with the gradient from unimodal loss. Besides, DGL removes the gradient back-propagated from the unimodal loss to the modality fusion module. This helps eliminate the gradient interference between the modality encoder and modality fusion module while ensuring their respective optimization processes. Finally, extensive experiments on multiple types of modalities, tasks, and frameworks with dense cross-modal interaction demonstrate the effectiveness and versatility of the proposed DGL. Code is available at  \href{https://github.com/shicaiwei123/ICCV2025-GDL}{https://github.com/shicaiwei123/ICCV2025-GDL}




\end{abstract}


\section{Introduction}

With the growing availability of affordable sensors, multimodal learning, which harnesses data from various sources, has received significant interest in machine learning. It is evident in various domains, including classification tasks~\cite{mm_cf1,mm_cf2,wei2023mshnet}, object detection~\cite{mm_detection1,mm_detection2,mm_detection3}, and segmentation tasks~\cite{rgbd_seg1,rgbd_seg2,rgbd_seg3}. Existing research on multimodal learning mainly focuses on developing fusion techniques, such as tensor-based fusion~\cite{tensor-1,tensor-2}, and attention-based fusion~\cite{mmformer,rfnet}. However, the simple combination of multiple modalities may not always yield satisfactory performance.

\begin{figure*}[t]
\centering
\includegraphics[width=1.0\textwidth]{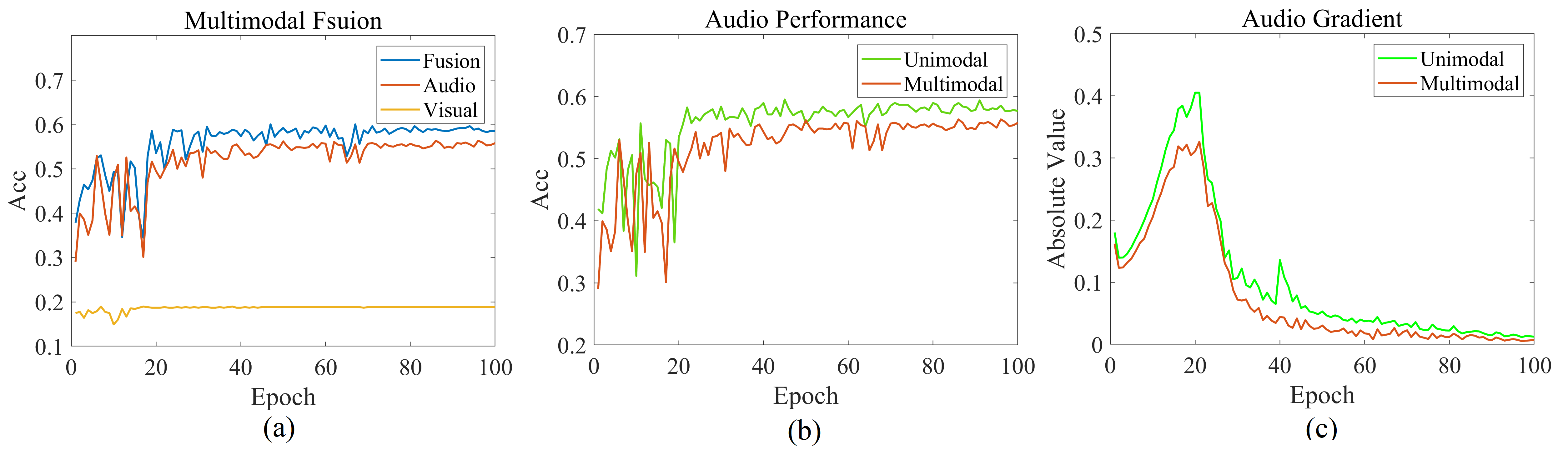} 
\caption{Visualization on the CREMA-D dataset. (a) illustrates the performance of each unimodal branch and their fusion in the multimodal model. (b) illustrates the performance of audio modality in the unimodal and multimodal models, respectively. (c) illustrates the gradient of audio modality back-propagated the unimodal and multimodal models, respectively.}
\label{diveristy-compare}
\vspace{-0.2cm}
\end{figure*}

Recent studies~\cite{ogm,umt} observe that the multimodal models could be inferior to the unimodal models in some situations. These studies attribute the decline in performance to imbalanced learning between modalities, in which the dominant modality that has better performance will suppress the optimization of the weaker modality, leading to inadequate feature learning in the weaker modality. To address this issue, techniques such as teacher aid~\cite{umt}, gradient modulation~\cite{ogm,agm,pmr,mmcosine}, and alternating optimization~\cite{mmpareto,diagnosing,MLA,reconboost} have been introduced to alleviate the imbalanced training and have shown promising results. However, these methods primarily focus on improving the weaker modality while overlooking the optimization of the dominant modality, which also exhibits sub-optimal performance compared to unimodal learning. As shown in Fig.~\ref{diveristy-compare}, the audio branch that has better performance in the multimodal model (see Fig.~\ref{diveristy-compare} (a)) also under-performs its unimodal baseline (see Fig.~\ref{diveristy-compare} (b)).



To this end, this paper focuses on the question of what makes the dominant modality under-optimized in multimodal learning. Here, we reveal that each modality in the multimodal model, even the highest-performing one, learns less effectively than in unimodal models due to optimization conflicts with the fusion module. Specifically, compared with the unimodal model, we prove that the modality fusion module in multimodal models will suppress the gradient back-propagated to the modality encoder. Moreover, the suppression degree will increase with the optimization progress. As shown in Fig.~\ref{diveristy-compare}(c), the gradient back-propagated to the audio encoder from the multimodal loss is smaller than that from the unimodal loss. And this gap becomes larger in the middle of the training. Consequently, the performance of each modality in the multimodal model is inferior to that in the unimodal model.



To tackle this issue, we introduce the disentangled gradient learning (DGL) strategy. Specifically, DGL first truncates the gradient from the modality fusion module propagating to the modality encoder. This helps avoid the problem of gradient suppression. Then DGL introduces independent unimodal loss for each modality encoder via the parameter-free modality dropout technique. This provides an independent gradient path for each modality encoder, eliminating inter-modality interference and enabling their optimization. Moreover, the gradient from the unimodal loss propagating to the modality fusion modules is also removed, avoiding the gradient interference between the unimodal loss and the multimodal loss. This enables the normal optimization of the fusion module in the multimodal model. Notably, the DGL relies solely on the representations of each modality, free from constraints of model structures and fusion methods, making it versatile for diverse scenarios. In summary, our contributions are as follows:

\begin{itemize}
  \item We reveal that the cross-modal interaction in multimodal learning decreases the gradient back-propagating to each modality encoder, leading to their inferior learning. This helps explain the insufficient optimization phenomenon of the dominant modality in multimodal learning.

  \item We introduce DGL to truncate the gradient from the modality fusion module propagating to the modality encoder and replace it with the gradient from independent unimodal loss. This helps eliminate optimization conflicts between the fusion module and encoders, as well as interference among different encoders.

  
  \item Extensive experiments demonstrate that 1) DGL can achieve considerable improvements over existing methods; 2) DGL is modality-data, fusion-method, and model-structure agnostic, offering high generality.

\end{itemize}

\section{Related Work}

\subsection{Multimodal Learning}

Multimodal learning utilizes complementary information contained in multimodal data to improve the performance of various tasks. One important direction in this area is the design of modality fusion methods, such as tensor-based fusion~\cite{tensor-1,tensor-2} and attention-based fusion~\cite{mmformer,rfnet}. Moreover, considerable efforts have been dedicated to harnessing information from multiple modalities to enhance model performance in specific tasks compared to unimodal frameworks. These tasks include action recognition~\cite{action_multi_all_1,MARS,MH2}, semantic segmentation~\cite{rgbd_seg1,rgbd_seg2,rgbd_seg3} and audio-visual speech recognition~\cite{AV1,AV2,AV3}. Yet, in joint training strategies, many multimodal methods often don't fully utilize all modalities, resulting in suboptimal unimodal representations. This leads to multimodal model performance falling short of expectations, even inferior to their unimodal counterparts.

\subsection{Under-optimization in Multimodal Learning}


Recent studies pointed out that most multimodal learning methods fail to enhance performance significantly even with more information~\cite{ogm,umt,wh,pmr,mmcosine,agm,MLA}. Wang~\etal~\cite{wh} observed that different modalities exhibit varying convergence rates, leading to multimodal models that fail to surpass their unimodal counterparts. Peng~\etal~\cite{ogm} further showed that the modality with superior performance tends to dominate the optimization process, leading to inadequate feature learning in weaker modalities. To this end, various methods have been developed to enhance the conventional multimodal learning framework and can be roughly categorized into two types: gradient modulation and alternating optimization.

Gradient modulation~\cite{ogm,pmr,mmcosine,agm} aims to enlarge the gradient of weaker modality in multimodal learning, balancing the optimization of different modality encoders. Specifically, OGM~\cite{ogm} proposes on-the-fly gradient modulation to manage the optimization of each modality adaptively. MMCosine~\cite{mmcosine} performs modality-wise L2 normalization to features and weights towards balanced and better multi-modal fine-grained learning. PMR~\cite{pmr} proposes the prototypical rebalancing strategy to hasten the learning of the slower modality and reduce the dominance of the stronger one. AGM~\cite{agm} introduces an adaptive gradient modulation method that can boost the performance of multimodal models with various fusion strategies. While gradient modulation methods show good results, improving weak modalities often degrades the performance of strong ones due to the inter-modality conflict.


To this end, the alternating optimization methods are proposed to improve the unimodal learning of each modality, including the domain one, in multimodal learning. MLA~\cite{MLA} transforms the conventional joint multimodal learning process into an alternating unimodal learning process to minimize inter-modality interference directly. ReconBoost~\cite{reconboost} updates a fixed modality each time via a dynamical learning objective to overcome the competition with the historical models. MMPareto~\cite{mmpareto} leverages the Pareto integration technique to catch innocent unimodal assistance, avoiding its conflict with multimodal optimization. Besides, Wei~\etal~\cite{diagnosing} proposes the Diagnosing \& Re-learning method to overcome the intrinsic limitation of modality capacity via the network re-initialization technique. While these methods improve the performance of each modality in multimodal learning simultaneously, they still fail to explain why the dominant modality in multimodal models underperforms that in unimodal learning.

\section{Method}
\label{MSD}

In this section, we analyze the under-optimization problem in the conventional multimodal learning paradigm, and then we describe the details of our proposed disentangled gradient learning framework.


\begin{figure*}[t]
\centering
\includegraphics[width=1.0\textwidth]{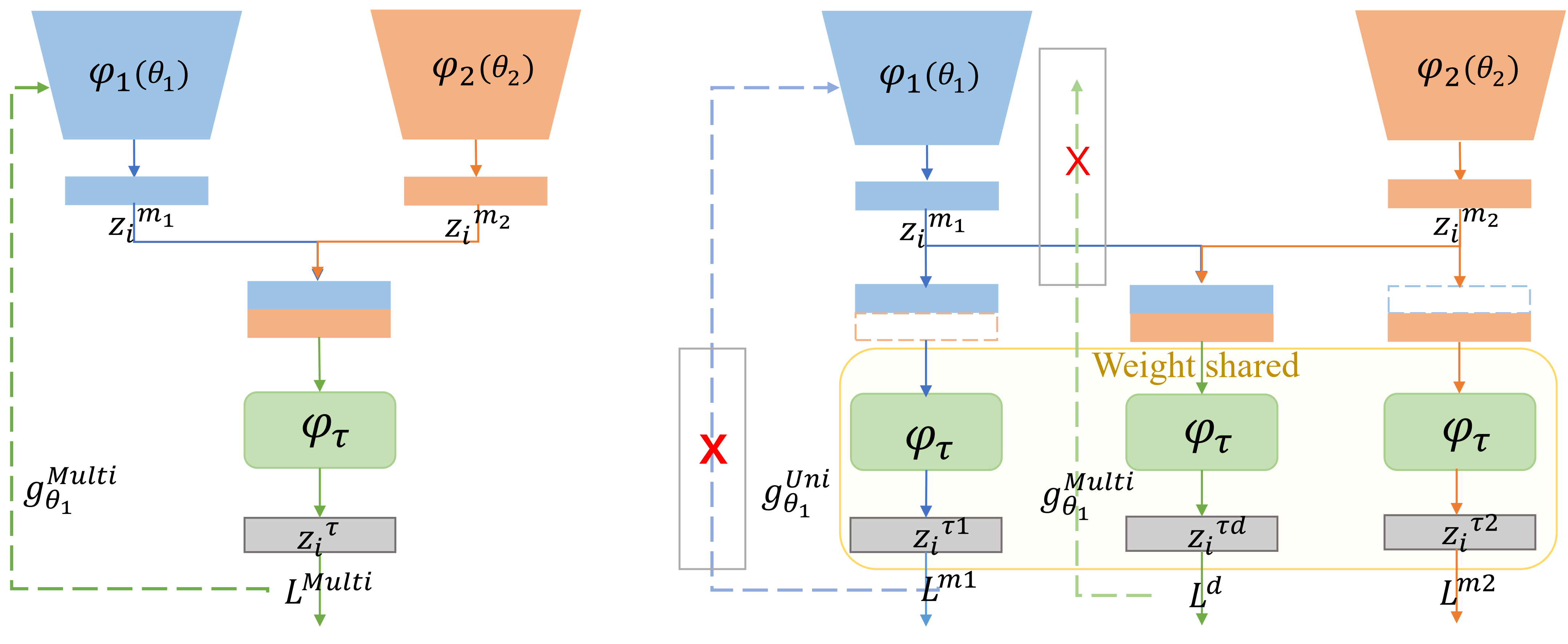} 
\caption{Architecture of vanilla multimodal model (a) and multimodal model with DGL (b). Compared with the vanilla model, DGL truncates the gradient back-propagated from the multimodal loss to the modality encoder and replaces it with the gradient from the unimodal loss, eliminating the gradient suppression on the modality encoder. Besides, DGL removes the gradient back-propagated from the unimodal loss to the modality fusion module, avoiding the gradient conflict between unimodal and multimodal losses. The red {\color{red}X} means the operation of gradient truncation; the block rectangle denotes the truncation region. }
\label{framework}

\end{figure*}


\subsection{Under-optimization Analysis in Multimodal Learning.} 

Existing methods attribute the insufficient multimodal performance to imbalanced learning, where the dominant modality will suppress the learning of the weak one. We reveal that all modality in the multimodal model, including the dominant modality, is under-optimized compared to the unimodal model due to the optimization conflict between the modality encoder and fusion module in the multimodal model.



Without loss of generality, we consider two input modalities as $m_1$ and $m_2$. As shown in Fig.~\ref{framework} (a), conventional multimodal learning can be described as follows: given a training set $\mathcal{D}=\left\{x_i, y_i\right\}_{i \in[N]}$, where the inputs $x_i=$ $\left(x_i^{m_1}, x_i^{m_2}\right)$ and $y_i \in[1,2,...,K]$, where K is the number of categories. We use two neural network encoders $\varphi_1\left(\theta_1, \cdot\right)$ and $\varphi_2\left(\theta_2, \cdot\right)$ to map each modality of the inputs to ${z_i}^{m_1} = \varphi_1\left(\theta_1, x_i^{m_1}\right) \in \mathbb{R}^{d_{1}}$ and ${z_i}^{m_2} = \varphi_2\left(\theta_2, x_i^{m_2}\right) \in \mathbb{R}^{d_{2}}$. Here $\theta_1, \theta_2$ are the parameters of $\varphi_1$ and $\varphi_2$ respectively. let $\varphi_{\tau}\left(\theta_{\tau}, \cdot,\cdot,\right)$ denote the fusion module, $W \in \mathbb{R}^{K \times\left(d_{1}+d_{2}\right)}$ and $b \in \mathbb{R}^K$ denote the parameters of the linear classifier to produce the logits output. The classification loss value of input $x_{i}$ in a multimodal model can be expressed as follows,
\begin{numcases}{}
\label{l1}
    L^{Multi}=L_{CE}(W \boldsymbol{z_i}^{\tau}+b,y_i) \\
    \label{zf}
{z_i}^{\tau}=\varphi_{\tau}(\theta_{\tau},{z_i}^{m_1},{z_i}^{m_2})
\end{numcases} where $L_{CE}(.,.)$ is the the cross-entropy loss function.

Then, according to the chain rule, the gradient $ g_{\theta_1}^{Multi}$ passed back to the encoder  $\varphi_1\left(\theta_1, \cdot\right)$ in the multimodal model can be expressed as follows,

\begin{equation}
\label{gradient_m1}
    g_{\theta_1}^{Multi}=\frac{\partial  L^{Multi}}{\partial f({z_i}^{\tau})} \frac{\partial f({z_i}^{\tau})}{\partial {z_i}^{\tau}} \frac{\partial {z_i}^{\tau}}{\partial {z_i}^{m_1}}  
\end{equation} where $f({z_i}^{\tau})=W \boldsymbol{z_i}^{\tau}+b$

Here, we set the fusion module as concatenation, which is the most widely used vanilla fusion method, thus  $\boldsymbol{z_i}^{\tau}=[\boldsymbol{z_i}^{m_1};\boldsymbol{z_i}^{m_2}]$. According to the gradient formula for the cross-entropy loss, we can rewrite $ g_{\theta_1}^{Multi}$ as follows,



\begin{equation}
\label{gradient_multi}
    g_{\theta_1}^{Multi}=(\frac{e^{(W_{y_{i}}^{m_{1}}z_{i}^{m_{1}}+b_1)}}{\sum_{k=1}^{K} e^{(W_{k}^{m_{1}}z_{1}^{m_{1}}+b_1)}e^{(W_{k}^{m_{2}}-W_{y_i}^{m_{2}}) z_{i}^{m_{2}} }}-1)  W_{y_{i}}^{m_{1}}
\end{equation} where $W=[W^{m_1},W^{m_2}]$, $W^{m_1}\in \mathbb{R}^{d_{1}}$,  $W^{m_2}\in \mathbb{R}^{d_{2}}$,$W_j$ denote the $j_{th}$ row of $W$.

Since the $y_{th}$ row of $W$ is the class center of the  $y_{th}$ class~\cite{center,center2}, the $z^{m_2}$ will gradually approach $W_{y_i}$ and move away from $W_{k}$ when optimizing the multimodal model. As a result, we can get the range of $e^{(W_{k}^{m_{2}}-W_{y_i}^{m_{2}}) z_{i}^{m_{2}} }$ as follow,

\begin{numcases}{}
\label{x1}
e^{(W_{k}^{m_{2}}-W_{y_i}^{m_{2}}) z_{i}^{m_{2}} }<1 , k\neq y_i\\
\label{x2}
    e^{(W_{k}^{m_{2}}-W_{y_i}^{m_{2}}) z_{i}^{m_{2}} }=1,   k= y_i
\end{numcases}

Besides, let $W=W^{m_1}$ and $z^{\tau}=z^{m_1}$, we can get the classification result of the vanilla unimodal model. The gradient $ g_{\theta_1}^{Uni}$ passed back to the encoder  $\varphi_1\left(\theta_1, \cdot\right)$ in the vanilla unimodal model can be expressed as follows,

\begin{equation}
\label{gradient_uni}
    g_{\theta_1}^{Uni}=(\frac{e^{(W_{y_{i}}^{m_{1}}z_{i}^{m_{1}}+b_1)}}{\sum_{k=1}^{K} e^{(W_{k}^{m_{1}}z_{1}^{m_{1}}+b_1)}}-1)  W_{y_{i}}^{m_{1}}
\end{equation}

According to Equation~\ref{x1} and~\ref{x2}, we can get the following inequality,

\begin{equation}
\label{grad_comp}
    abs(g_{\theta_1}^{Uni})>abs(g_{\theta_1}^{Multi})>0
\end{equation}

Because the parameters are updated along the negative gradient direction, the encoder $\varphi_1\left(\theta_1, \cdot\right)$ in the unimodal model will converge faster than that in the multimodal model. In other words, compared with unimodal models, multimodal models limit the optimization of modality encoders. Thus, each modality, including the dominant modality that has higher performance, in the multimodal model will suffer insufficient learning and has worse performance than that in the unimodal model (see Fig.~\ref{diveristy-compare} (b)). In addition, if $m_2$ is an easy-to-learn modality, $W_{y_i}^{m_2} z_{i}^{m_2}$ will be larger than $W_{y_i}^{m_1} z_{i}^{m_1}$, so $e^{(W_{k}^{m_{2}}-W_{y_i}^{m_{2}}) z_{i}^{m_{2}} }$ will be smaller than $e^{(W_{k}^{m_{1}}-W_{y_i}^{m_{1}}) z_{i}^{m_{1}} }$, which means that the gradient suppression caused by the easy-to-learn modality  will be greater than that of hard-to-learn one. This explains why the easy-to-learn modality performs better than the hard-to-learn one when the modality gradients interfere with each other.

More importantly, even if the learning difficulty of $m_1$ and $m_2$ is the same, the optimization constraints on each modality encoder will also become increasingly severe when the multimodal model is optimized (see Fig.~\ref{diveristy-compare} (c)). This is because $W_{y_i}^{m_2} z_{i}^{m_2}$ and  $W_{y_i}^{m_1} z_{i}^{m_1}$ becomes increasingly close to $1$ and the absolute value of $g_{\theta_1}^{Multi}$ becomes increasingly smaller than that of $g_{\theta_1}^{Uni}$.

To ensure the optimization of modality encoders in the multimodal model, we should constraint $e^{(W_{k}^{m_{2}}-W_{y_i}^{m_{2}}) z_{i}^{m_{2}} }=1$ so that $g_{\theta_1}^{Multi}=g_{\theta_1}^{Uni}$. Since each row of $W$ is the class center vector of different types, the cross entropy loss will constrain them to stay away from each other to achieve good discrimination. Thus, the distance between $(W_{k}^{m_{2}}$ and $W_{y_i}^{m_{2}}$ will always be larger than $0$. Consequently, the only way to ensure $e^{W_{k}^{m_{2}}-W_{y_i}^{m_{2}}) z_{i}^{m_{2}} }=1$ is to constrain ${z_i}^{m_2}=0$. Nevertheless, this indicates that the fusion module entirely neglects the data from the modality $m_2$, which deviates from the purpose of multimodal fusion to integrate information from multiple sources. Therefore, there is an optimization conflict between the modality encoder and modality fusion module, which limits the learning of modality encoders.

\begin{algorithm}[t]
    \caption{Multimodal learning with DGL strategy} 
    
    \label{dgd} 
    \begin{algorithmic}
        \REQUIRE Training dataset D, iteration number T, hyper-parameter $\alpha$.
        \STATE \quad \textbf{for} t = 0,..., T - 1 \textbf{do}
        \STATE \quad\quad Feed-forward the batched data to the model.
        \STATE \quad\quad Calculate unimodal loss $L^{m_1}$ and  $L^{m_2}$ via
        \STATE \quad\quad  Equation~\ref{x3} and ~\ref{x4}.
        \STATE \quad\quad  Calculate unimodal gradient.
        \STATE \quad\quad  Eliminate the gradient passed back from the
         \STATE \quad\quad  unimodal loss to the fusion module.
        \STATE \quad\quad Calculate multimodal loss $L^{d}$ with detach
        \STATE \quad\quad representation via Equation~\ref{ld}.        
        \STATE \quad \quad Calculate multimodal gradient.
        \STATE \quad\quad Update the model parameters.
    \end{algorithmic} 
\end{algorithm}

\subsection{Disentangled Gradient Learning} 
\label{pee}

As discussed, compared with unimodal models, multimodal models constrain the optimization of modality encoders via $e^{(W_{k}^{m_{2}}-W_{y_i}^{m_{2}}) z_{i}^{m_{2}} }$. To this end, we introduce DGL to decouple the optimization process between the modality encoder and the modality fusion module in the multimodal model by detaching and reorganizing their gradient propagation paths, as shown in Fig.~\ref{framework} (b). Compared with the vanilla model, this can eliminate the optimization constraint on modality encoders completely while ensuring the optimization of the modality encoder and fusion module. Specifically, DGL consists of two parts: gradient detaching to truncate the gradient back-propagation of multimodal loss and gradient reorganizing to boost the optimization of each modality encoder. 



\textbf{Gradient Detaching.} This stage aims to truncate the gradient $g_{\theta_1}^{Multi}$ transmitted from the modality fusion module to the modality encoder so that eliminates the suppression on the modality encoder optimization. To this end, we rewrite the Equation~\ref{zf} as follows,

\begin{numcases}{}
{z_i}^{\tau d}=\varphi_{\tau}(\theta_{\tau},{z_i}^{m_1}.detach(),{z_i}^{m_2}.detach())
\end{numcases} where `.detach()' is the PyTorch function to truncate the gradient. Then we calculate the multimodal loss as follows,

\begin{equation}
\label{ld}
    L^{d}=L_{CE}(W({z_i}^{\tau d})+b,y_i)
\end{equation}

Compared with vanilla loss $ L^{Multi}$ in Equation~\ref{l1}, $g_{\theta_1}^{Multi}$ calculated by the multimodal loss $ L^{d}$ will not be passed back to encoder $\varphi_1\left(\theta_1, \cdot\right)$, and thus will not affect its parameter optimization.

\textbf{Gradient Reorganizing.} Although gradient detaching eliminates the constraint on the modality encoder optimization, it also blocks the optimization of encoder$\varphi_1\left(\theta_1, \cdot\right)$ because no gradient can be passed back to the encoder. To address this problem, we consider introducing unimodal gradients for each encoder to optimize their parameters.

Unlike existing methods that introduce extra classifiers for modality representations, the proposed DGL calculates the unimodal loss via the modality dropout technique~\cite{wei2023mmanet,wei2025robust}. This retains the fusion module and calculates the unimodal loss by setting other modality representations as $0$ directly. This enables DGL to handle dense cross-modal interaction. More importantly, the module parameters are shared with unimodal and multimodal loss, introducing no extra structure complexity. 

Specifically, take encoder $\varphi_1\left(\theta_1, \cdot\right)$ as an example, its unimodal loss is defined as follows,

\begin{numcases}{}
\label{x3}
    L^{m_1}=L_{CE}(W({z_i}^{\tau 1})+b,y_i)\\
    {z_i}^{\tau 1}=\varphi_{\tau}(\theta_{\tau},{z_i}^{m_1},0)
\end{numcases} where the parameters $\theta_{\tau}, W$ and $b$ are shared with those in  $L^{d}$.  

According to the chain rule, the gradient passed back to the encoder $\varphi_1\left(\theta_1, \cdot\right)$ is exactly the vanilla $g_{\theta_1}^{Uni}$ defined in Equation~\ref{gradient_uni}. Similarly, we can get the unimodal loss for encoder $\varphi_2\left(\theta_2, \cdot\right)$ as follows,

\begin{numcases}{}
\label{x4}
    L^{m_2}=L_{CE}(W({z_i}^{\tau 2})+b,y_i)\\
    {z_i}^{\tau 2}=\varphi_{\tau}(\theta_{\tau},0,{z_i}^{m_2})
\end{numcases}

It is worth mentioning that the gradient of the unimodal loss also passes through the fusion module.  To avoid conflicting with the multimodal gradient, we will set the unimodal gradient in the fusion module to 0 before calculating the gradient of the multimodal loss, as described in Algorithm~\ref{dgd}.

So far, we have successfully eliminated the optimization conflict between the modality encoder and the modality fusion module when ensuring their respective optimization processes. More importantly, since the extra unimodal gradient paths for each modality encoder are independent, DGL also addresses the inter-modality competition that limits the performance of multimodal model~\cite{ogm,pmr}.

\textbf{Total loss.} As discussed above, the total loss $L_{DGL}$ for multimodal learning with disentangled gradient learning is defined as follows,

\begin{equation}
    L_{DGL}=[L^{d},\alpha(L^{m_1} + L^{m_2})]
\end{equation} where $L^{d}$ is the multimodal loss with gradient detaching. $L^{m_1}$ and $L^{m_2}$ are unimodal loss. Here we use the operation of concatenation instead of summation since the gradient of $L^{d}$ and $L^{m_1} + L^{m_2}$ is decoupled, as described in Algorithm~\ref{dgd}. Specifically, $L^{d}$ only optimizes the modality fusion module and the fully connected layer in the multimodal model. $L^{m_1} + L^{m_2}$ only optimizes the modality encoders in the multimodal model. $\alpha$ is the hyper-parameter to calibrate the optimization of the modality encoder, which has been widely proven to improve the modality capacity and multimodal performance.

\begin{table*}[]
\centering
\renewcommand\arraystretch{1.0}
\setlength{\tabcolsep}{3mm}{
\begin{tabular}{c|ccc|ccc|ccc}
\hline
\multirow{2}{*}{Method} & \multicolumn{3}{c|}{CREMA-D}                                                                           & \multicolumn{3}{c|}{KS}                                                                                & \multicolumn{3}{c}{VGGSound}                                                                          \\ \cline{2-10} 
                        & \multicolumn{1}{c}{Audio}          & \multicolumn{1}{c}{Visual}         & Multi                      & \multicolumn{1}{c}{Audio}          & \multicolumn{1}{c}{Visual}         & Multi                      & \multicolumn{1}{c}{Audio}          & \multicolumn{1}{c}{Visual}         & Multi                      \\ \hline
Audio-only              & \multicolumn{1}{c}{62.18}          & \multicolumn{1}{c}{-}               &      -                      & \multicolumn{1}{c}{48.7}           & \multicolumn{1}{c}{-}               &                -            & \multicolumn{1}{c}{45.13}          & \multicolumn{1}{c}{-}               &    -                        \\ 
Visual-only             & \multicolumn{1}{c}{-}               & \multicolumn{1}{c}{68.23}          &           -                 & \multicolumn{1}{c}{-}               & \multicolumn{1}{c}{54.6}           &                -            & \multicolumn{1}{c}{-}               & \multicolumn{1}{c}{30.68}          &                 -           \\ 
Concat                  & \multicolumn{1}{c}{59.62}          & \multicolumn{1}{c}{36.71}          & 65.1                      & \multicolumn{1}{c}{43.35}           & \multicolumn{1}{c}{48.72}          & 64.45                      & \multicolumn{1}{c}{43.18}          & \multicolumn{1}{c}{20.55}          & 47.90                       \\ \hline
G-Blending              & \multicolumn{1}{c}{58.78}          & \multicolumn{1}{c}{58.62}          & 69.21                      & \multicolumn{1}{c}{46.35}          & \multicolumn{1}{c}{51.12}          & 69.60                      & \multicolumn{1}{c}{44.34}          & \multicolumn{1}{c}{26.56}          & 49.33                      \\ 
OGM\_GE                  & \multicolumn{1}{c}{57.76}          & \multicolumn{1}{c}{40.09}          & 68.82                      & \multicolumn{1}{c}{44.23}          & \multicolumn{1}{c}{45.81}          & 66.89                      & \multicolumn{1}{c}{41.85}          & \multicolumn{1}{c}{27.41}          & 48.71                      \\ 
PMR                     & \multicolumn{1}{c}{55.11}          & \multicolumn{1}{c}{38.34}          & 67.44                      & \multicolumn{1}{c}{43.61}          & \multicolumn{1}{c}{46.67}          & 65.70                      & \multicolumn{1}{c}{42.33}          & \multicolumn{1}{c}{25.12}          & 48.17                      \\ 
AGM                     & \multicolumn{1}{c}{56.37}          & \multicolumn{1}{c}{43.54}          & \multicolumn{1}{c|}{69.61} & \multicolumn{1}{c}{46.12}          & \multicolumn{1}{c}{47.65}          & \multicolumn{1}{c|}{68.88} & \multicolumn{1}{c}{41.87}          & \multicolumn{1}{c}{27.34}          & \multicolumn{1}{c}{49.10} \\ 
MLA                     & \multicolumn{1}{c}{60.46}          & \multicolumn{1}{c}{64.23}          & 73.12                      & \multicolumn{1}{c}{\underline{50.03}}          & \multicolumn{1}{c}{\underline{54.67}}          & \underline{71.12}                      & \multicolumn{1}{c}{\underline{45.87}}          & \multicolumn{1}{c}{\underline{31.60}}          & \underline{51.19}                      \\ 
MMPareto                & \multicolumn{1}{c}{59.43}          & \multicolumn{1}{c}{61.09}          & 70.12                      & \multicolumn{1}{c}{48.40}          & \multicolumn{1}{c}{52.42}          & 69.83                      & \multicolumn{1}{c}{42.44}          & \multicolumn{1}{c}{30.07}          & 49.12                      \\ 
D\&R                     & \multicolumn{1}{c}{\underline{61.11}}           & \multicolumn{1}{c}{\underline{64.57}}          & \underline{74.32}                       & \multicolumn{1}{c}{49.78}          & \multicolumn{1}{c}{54.88}          & 69.10                      & \multicolumn{1}{c}{45.18}          & \multicolumn{1}{c}{31.23}          & 50.84                      \\ \hline
DGL                     & \multicolumn{1}{c}{\textbf{63.12}} & \multicolumn{1}{c}{\textbf{69.11}} & \textbf{77.48}             & \multicolumn{1}{c}{\textbf{52.89}} & \multicolumn{1}{c}{\textbf{60.11}} & \textbf{74.78}             & \multicolumn{1}{c}{\textbf{47.13}} & \multicolumn{1}{c}{\textbf{33.45}} & \textbf{52.53}             \\ \hline
\end{tabular}}
\caption{Comparison with existing modulation strategies on CREMA-D, Kinetics-Sounds, and VGGSound datasets. Bold and underline mean the best and second-best results, respectively. The proposed DGL achieves the best performance in both unimodal and multimodal performance comparisons. The metric is accuracy.}
\label{compare}
\vspace{-1.0em}
\end{table*}
%

\subsection{Relationship to Prior Work} 

Some concurrent works~\cite{wh,mmpareto} seem similar to our DGL, which also introduces unimodal loss to assist multimodal learning. However, there are still differences in motivations, implementations, and performance with ours. They focus on the optimization conflict between different modalities and leverage the unimodal loss to balance their optimization progress. Here, the unimodal loss and multimodal loss will be used to update all parameters simultaneously. In contrast, we address the optimization conflict between the modality encoder and the modality fusion module. And the unimodal model loss in DGL is only used to update the modality encoder, while multimodal loss is only used to update the fusion module and classifier, eliminating the interference between them. As a result, DGL achieves better performance in unimodal and multimodal evaluation than existing unimodal assistance methods (see Tables~\ref{compare} and ~\ref{gene}).

Besides, existing methods usually introduce extra classifiers for modality representations In contrast, the proposed DGL overcomes those defects by retaining the fusion module and setting other modality representations as $0$ to calculate the unimodal directly. 



\section{Experiments}

\subsection{Datasets} 

\textbf{CREMA-D}~\cite{Crema} serves as an audio-visual dataset crafted for speech emotion recognition. It contains 7442 video clips of 2-3 seconds from 91 actors for 6 emotions. The whole dataset is randomly divided into 6698  samples as the training set and 744 samples as the testing set.


\textbf{Kinetics-Sounds (KS)}~\cite{ks} is a dataset formed by filtering the Kinetics dataset for 34 human action classes which have been chosen to be potentially manifested visually and aurally. This dataset contains 19k 10-second video clips (15k training, 1.9k validation, 1.9k test).

\textbf{VGGSound}~\cite{vggsound} is a large-scale video dataset with 309 categories, capturing various audio events in daily life. For our experiment, we employed 168,618 videos for training and validation, and 13,954 videos for testing.

\textbf{MOSI}~\cite{mosi} is a popular Audio-Visual-Text dataset for the multimodal sentiment analysis task. It collects 2,199 utterance-video clips of 93 monologue videos from YouTube, each of which is labeled with a continuous sentiment score ranging from -3 (strongly negative) to 3 (strongly positive). In this paper, we use the two-class label setting that considers positive/negative results only. We use this dataset to prove the method’s generalization to the scene with multiple modalities.

\textbf{NYUv2}~\cite{nyuv2} is a RGB-D dataset for indoor semantic segmentation task. It comprises 1,449 indoor RGB-D images, of which 795 are used for training and 654 for testing. We used the common 40-class label setting. We use this dataset to prove the method’s effectiveness beyond the audio-visual dataset and classification task.

\subsection{Comparison on the multimodal task}

\textbf{Implementation details.} For a fair comparison, we adopt the same setting as the previous method~\cite{ogm} for the audio-visual task. For datasets CREMA-D, KS, and VGGSound, we adopt the ResNet18 as the encoder backbone, mapping the input data to 512-dimensional vectors.  For CREMA-D, we extract 1 frame from each of the clips as the visual input. The whole audio data is transformed into a spectrogram of size 257×299 by librosa~\cite{librosa} using a window with a length of 512 and an overlap of 353. For the KS and VGGSound datasets, we extract 3 frames from each clip as visual inputs as visual input and process audio data into a spectrogram of size 257×1004. We use SGD with 0.9 momentum and 1e-4 weight decay as the optimizer. The learning rate is 2e-3 initially and multiplies 0.1 every 70 epochs. The training batch size and epoch are 100 and 64, respectively. 


\textbf{Comparison settings.} To study the advantage of DGL, we make comparisons with three modulation approaches, OGM~\cite{ogm}, AGM~\cite{agm}, PMR~\cite{pmr}, and four alternating optimization methods, G-Blending~\cite{wh}, MLA~\cite{MLA}, MMPareto~\cite{mmpareto} and D\&R~\cite{diagnosing}. For a fair comparison, we unify the backbone as ResNet18 and the fusion method as concatenation in all experiments.  Note that the original MLA uses epoch as 150 and batch size as 16. For a fair comparison, we unify them with other comparison methods as epoch 100 and batch size 64.


\begin{table*}[]
\centering
\setlength{\tabcolsep}{3mm}{
\begin{tabular}{c|ccc|cccc}
\hline
\multirow{3}{*}{Method} & \multicolumn{3}{c|}{CREMA-D}                                                               & \multicolumn{4}{c}{MOSI}                                                                                                        \\ \cline{2-8} 
                        & \multicolumn{1}{c|}{Audio}          & \multicolumn{1}{c|}{Visual}         & Multi          & \multicolumn{1}{c|}{Audio}          & \multicolumn{1}{c|}{Visual}         & \multicolumn{1}{c|}{Text}           & Multi          \\ \cline{2-8} 
                        & \multicolumn{3}{c|}{MMTM-Fusion}                                                           & \multicolumn{4}{c}{MLP-Fusion}                                                                                                  \\ \hline
Audio-only              & \multicolumn{1}{c}{59.18}          & \multicolumn{1}{c}{-}              & -              & \multicolumn{1}{c}{44.25}          & \multicolumn{1}{c}{-}              & \multicolumn{1}{c}{-}              & -              \\ 
Visual-only             & \multicolumn{1}{c}{}               & \multicolumn{1}{c}{68.34}          &                & \multicolumn{1}{c}{-}              & \multicolumn{1}{c}{54.28}          & \multicolumn{1}{c}{-}              & -              \\ 
Text-only               & \multicolumn{1}{c}{-}              & \multicolumn{1}{c}{-}              & -              & \multicolumn{1}{c}{}               & \multicolumn{1}{c}{-}              & \multicolumn{1}{c}{77.12}          & -              \\ \hline
Multimodal Baseline      & \multicolumn{1}{c}{56.11}          & \multicolumn{1}{c}{23.43}          & 60.12          & \multicolumn{1}{c}{40.65}          & \multicolumn{1}{c}{51.33}          & \multicolumn{1}{c}{76.33}          & 76.83          \\ 
G-Blending                      & \multicolumn{1}{c}{57.72}          & \multicolumn{1}{c}{62.11}          & 69.99          & \multicolumn{1}{c}{42.39}          & \multicolumn{1}{c}{53.03}          & \multicolumn{1}{c}{76.97}          & 78.04          \\ 
OGM\_GE                     & \multicolumn{1}{c}{55.61}          & \multicolumn{1}{c}{40.61}          & 65.31          & \multicolumn{1}{c}{42.09}          & \multicolumn{1}{c}{52.68}          & \multicolumn{1}{c}{75.87}          & 77.97          \\ 
PMR                     & \multicolumn{1}{c}{54.71}          & \multicolumn{1}{c}{39.21}          & 64.89          & \multicolumn{1}{c}{41.89}          & \multicolumn{1}{c}{52.43}          & \multicolumn{1}{c}{75.67}          & 77.54          \\ 
AGM                     & \multicolumn{1}{c}{55.29}          & \multicolumn{1}{c}{44.18}          & 66.65         & \multicolumn{1}{c}{42.24}          & \multicolumn{1}{c}{52.97}          & \multicolumn{1}{c}{76.04}          & 77.95          \\ 
MMPareto                & \multicolumn{1}{c}{56.79}          & \multicolumn{1}{c}{63.21}          & 69.88          & \multicolumn{1}{c}{42.92}          & \multicolumn{1}{c}{53.08}          & \multicolumn{1}{c}{76.87}          & 78.15          \\ 
D\&R                     & \multicolumn{1}{c}{\underline{58.67}}          & \multicolumn{1}{c}{\underline{66.32}}          & \underline{72.28}          & \multicolumn{1}{c}{\underline{43.65}}          & \multicolumn{1}{c}{\underline{53.81}}          & \multicolumn{1}{c}{\underline{77.01}}          & \underline{78.81}          \\ \hline
DGL                     & \multicolumn{1}{c}{\textbf{60.30}} & \multicolumn{1}{c}{\textbf{69.50}} & \textbf{75.00} & \multicolumn{1}{c}{\textbf{44.78}} & \multicolumn{1}{c}{\textbf{55.11}} & \multicolumn{1}{c}{\textbf{77.39}} & \textbf{79.78} \\ \hline
\end{tabular}}
\caption{\textbf{Left}: Comparison with imbalanced multimodal learning methods with dense cross-modal interaction on the CREMA-D dataset. \textbf{Right}: Comparison with imbalanced multimodal learning methods on the MOSI dataset with three modalities. }
\label{gene}
\end{table*}

\textbf{Results}. As shown in Table~\ref{compare}, the proposed DGL framework achieves the best performance on all datasets with a significant improvement compared to existing methods. Compared to the second-best methods, MLA and D\&R, it improves the multimodal performance by 3.15\%, 3.66\%,  and 1.34\% on CREMA-D, KS, and VGGSound datasets, respectively. These results demonstrate the superiority of the proposed DGL technique. 

More importantly, the unimodal performance in the multimodal model trained with the DGL also outperforms that in the unimodal model. This verifies the effectiveness of DGL in promoting unimodal learning in multimodal learning. As a comparison, while existing gradient modulation methods, such as OGM\_GE, can boost the weaker visual modality's performance in CREMA-D above 36.71\% compared to vanilla concatenation fusion, they also reduce the dominant audio modality’s performance below 59.62\%. Then, the alternating optimization methods introduce unimodal assistance to improve the performance of audio and visual modalities in the multimodal model simultaneously. However, they overlook the optimization conflicts between the fusion module and encoders, failing to improve the unimodal performance in the multimodal model to be similar as that in the unimodal model. In contrast, DGL addresses this defect by truncating the gradient from the modality fusion module propagating to the modality encoder. As a result, the performance of the unimodal branch within the multimodal model trained with DGL surpasses that of the vanilla unimodal model.

\subsection{Ablation Study}

We first conduct experiments to discuss the influence of gradient truncation and hyper-parameters $\alpha$, then we study the generalization of DGL to different fusion methods, modality types, modality numbers, and tasks. Limited by the page, more ablation experiments and discussion can be seen in the supplementary materials.

\textbf{The effect of gradient truncation.} We conduct experiments on the CREMA-D dataset to study the effectiveness of multimodal gradient truncation (MT) and unimodal gradient truncation(UT). As shown in Table~\ref{ut}, both MT and UT improve the performance significantly compared to the baseline model. The former avoids the gradient suppression from multimodal fusion to the unimodal encoder, and the latter avoids the optimization conflict between multimodal and unimodal loss. Moreover, their combination can achieve better performance.

\begin{table}[t]
\centering

\begin{tabular}{c|cccc}
\hline
Setting & Baseline & +MT & +UT & DGL(MT+UT)   \\ \hline
ACC     & 65.10    & 73.31                  & 74.22                & \textbf{77.48} \\ \hline
\end{tabular}
\caption{Ablation results of gradient truncation on the CREMA-D.}
\label{ut}

\end{table}



\begin{table}[]
\centering
\begin{tabular}{c|ccccc}
\hline
$\alpha$ & 1     & 2     & 3     & 4     & 5     \\ \hline
G-Blending                    & 67.89 & 68.12 & 69.20 & \textbf{69.21} & 68.67  \\ 
DGL                   & 73.35 & 75.46 & 76.13 & \textbf{77.48} & 77.08 \\ \hline
\end{tabular}
\caption{Results of the DGL and G-Blending with different $\alpha$ on CREMA-D. The fusion strategy is concatenation.}
\label{alpha}
\vspace{-0.2cm}
\end{table}

\textbf{The effect of $\alpha$}. We investigate the influence of different $\alpha$ for DGL on the CREMA-D dataset. As shown in Table~\ref{alpha}, the performance of DGL increases with parameter $\alpha$, indicating that enhancing the optimization of unimodal learning can improve multimodal performance, which is consistent with the conclusion of existing methods~\cite{wh,mmpareto}. However, although existing methods, such as G-Blending, also introduce unimodal loss to boost unimodal learning in the multimodal model, their performance improvement is limited. This is because they ignore the optimization conflict between the modality encoder and the modality fusion module. In contrast, DGL eliminates this conflict and achieves significant improvement.

\textbf{Generalization to dense cross-modal interaction.} Existing experiments are conducted with vanilla concatenation fusion. To validate the applicability of DGL in the multimodal model with dense cross-modal interaction, we apply it to two intermediate fusion methods MMTM~\cite{mmtm} and MLP-Mixer~\cite{cube} for CREMA-D and MOSI datasets. Specifically, MMTM is a CNN-based architecture that fuses intermediate feature maps from different modalities via multimodal squeeze and excitation operations, while MLP-Mixer is an MLP-based architecture that performs fusion using affine transformations on intermediate feature maps. For both methods, we use only one frame for each video on each dataset to align their input on the benchmark. 

Since MLA~\cite{MLA} only works with parameter-shared fusion, it is excluded from this experiment. For a fair comparison, we set the modality representation to $0$ to obtain unimodal logits for G-Blending, OGM\_GE, and MMPareto. As shown in Table~\ref{gene}, DGL consistently surpasses the multimodal baseline and other methods in both multimodal and unimodal performance. Additionally, the unimodal performance of the DGL multimodal model also exceeds that of the unimodal model. These results confirm the applicability of DGL in scenarios with dense cross-modal interaction.

\textbf{Generalization to the multiple-modality case.} Existing methods mainly only focus on the case of two modalities~\cite{ogm,agm,pmr}, limiting their applicability to broader, more complex scenarios. In contrast, DGL imposes no restrictions on the number of modalities, allowing for greater flexibility. Here, we conduct experiments on the MOSI dataset with three modalities: audio, vision, and text. For a comprehensive comparison, we retain the core uni-modal balancing strategy of G-Blending, OGM\_GE, AGM, PMR, and MMPareto and extend them to more than two modality cases. As shown in the left section of Table~\ref{gene}, DGL achieves the best performance in multimodal and unimodal performance,  demonstrating its flexibility in such scenarios.

\begin{table}[t]
\centering

\begin{tabular}{cc|cc}
\hline
\multicolumn{2}{c|}{\multirow{2}{*}{Initialization}} & \multicolumn{2}{c}{Method}                                     \\ \cline{3-4} 
\multicolumn{2}{c|}{}                                & \multicolumn{1}{c}{ESANet}   & DGL   \\ \hline
\multicolumn{2}{c|}{From Scratch}                    & \multicolumn{1}{c}{38.59} & \textbf{41.67} \\ 
\multicolumn{2}{c|}{ImageNet Pre-train}              & \multicolumn{1}{c}{48.48}  & \textbf{50.10} \\ \hline
\end{tabular}
\caption{Model performance comparison of RGB-Depth semantic segmentation task on NYUv2 dataset. The metric is mIOU.}
\label{seg}
\vspace{-0.2cm}
\end{table}

\textbf{Generalization to the dense prediction task.} Existing methods require unimodal logit output or cluster center to calculate the modulation coefficients of different modalities. This limits their application in dense prediction scenarios. In contrast, DGL does not need these prerequisites, so it can be extended to non-classification tasks.

 We take ESANet as the baseline, which is an efficient and robust model for RGB-D segmentation. We keep all hyper-parameters the same as the official implementation. Specifically, the model is optimized by Adam for 300 epochs with a mini-batch 8 and learning rate 1e-2.
 

 As shown in Table~\ref{seg}, we train the network on both scenarios: (1) training from scratch on NYUv2; (2) pre-training on ImageNet followed by fine-tuning on NYUv2. For both scenarios, the proposed DGL significantly enhances the performance of ESANet. This confirms the ability of DGL to handle tasks beyond the audio-visual data and classification.

\begin{figure}[t]
\centering
\includegraphics[width=0.5\textwidth]{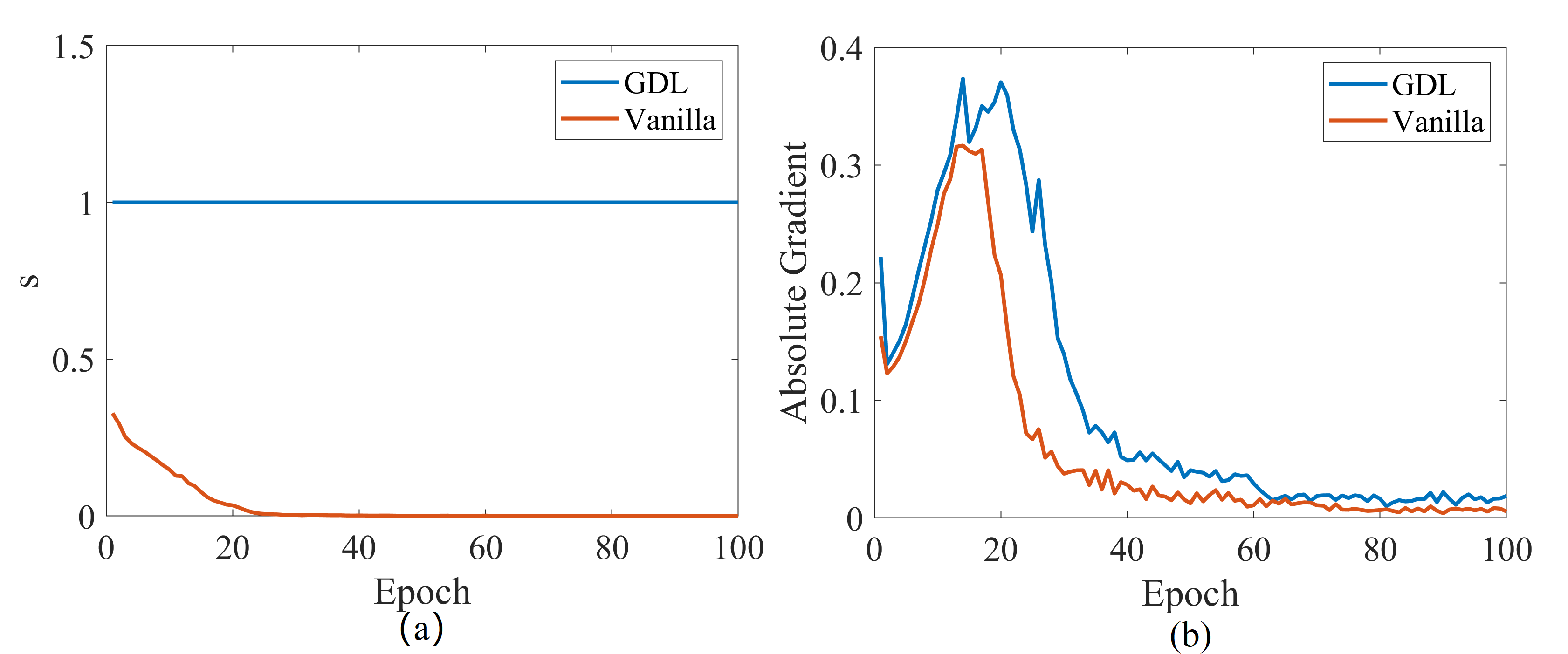} 
\caption{Visualization on the CREMA-D dataset. (a) illustrates $s=e^{(W_{k}^{m_{2}}-W_{y_i}^{m_{2}}) z_{i}^{m_{2}}}$ for the audio branch in the vanilla and DGL multimodal model.  (b) illustrates the gradient of the audio branch in the vanilla and DGL multimodal model.}
\label{audio-ana}
\end{figure}

\textbf{Visualization.} To further explain the mechanism of DGL, we visualize $s=e^{(W_{k}^{m_{2}}-W_{y_i}^{m_{2}}) z_{i}^{m_{2}}}$ and the training gradient for the audio branch in the vanilla and DGL multimodal model. As shown in Fig.~\ref{audio-ana}(a), $s$ of the vanilla model is always less than 1 and is close to 0 after the 20th epochs. As shown in Fig.~\ref{audio-ana}(b), the gradient from vanilla multimodal back-propagated to the audio encoder also decreases rapidly after the 20th epoch. This is consistent with the derivation of Equation~\ref{gradient_m1}. On the contrary, DGL mitigates gradient suppression caused by $s<1$ and maintains a large gradient to update the audio encoder after the 20th epoch. Therefore, as shown in Table~\ref{compare} and~\ref{gene}, the audio branch in the DGL model outperforms that in the vanilla model significantly. This demonstrates the effectiveness of disentangled gradient learning.

\section{Conclusion}
\label{cl}

In this study, we reveal the optimization conflict between the modality encoder and modality fusion module in the multimodal model. The cross-modal interaction will suppress the gradient back-propagated to each modality encoder and limit their optimization, including the dominant modality. To address this problem, we introduce DGL to decouple the optimization of the modality encoder and modality fusion module. First, DGL eliminates gradient suppression by truncating the gradient from the multimodal loss to the modality encoder and substituting it with the gradient from the unimodal loss. Additionally, DGL blocks the gradient from the unimodal loss to the modality fusion module, ensuring independent optimization of the modality encoder and fusion module.  Extensive experiments show that DGL can achieve consistent performance gain on multimodal classification and segmentation tasks with different model structures and fusion methods. 

{
    \small
    \bibliographystyle{ieeenat_fullname}
    \bibliography{main}
}

\end{document}